\documentclass{article}
\usepackage{graphicx}
\usepackage{hyperref} 
\usepackage{float}

\graphicspath{{figures/}{pictures/}{images/}{./}} 

\usepackage{times}                     

\usepackage{tabu}                      
\usepackage{booktabs}                  
\usepackage{lipsum}                    
\usepackage{mwe}                       
\usepackage{graphicx}
\usepackage{amsmath}
\usepackage[ruled,vlined]{algorithm2e}
\usepackage{mathptmx}                  
\usepackage{amssymb}
\usepackage{tikz}
\usepackage{tikz}
\usetikzlibrary{shapes.geometric, arrows.meta, positioning}

\tikzstyle{step} = [rectangle, rounded corners, minimum width=2.8cm,
                    minimum height=1.2cm, text centered, draw=black, fill=gray!15]
\tikzstyle{arrow} = [thick,->,>=stealth]

\title{Topological Invariant-Based Iris Identification via Digital Homology and Machine Learning}

\author{
Ahmet Öztel\thanks{e-mail: aoztel@bartin.edu.tr}\\ %
\parbox{1.6in}{\scriptsize \centering Department of Business Administration, Bartin University, Turkiye \\ PhD Student, Department of Mathematics, Ege University} %
\and
İsmet Karaca\thanks{e-mail: ismet.karaca@ege.edu.tr}\\ %
\scriptsize Department of Mathematics, Ege University, Turkiye
}

\begin{document}

\maketitle

\begin{figure}[h]
    \centering
    \makebox[0pt][c]{\includegraphics[height=3cm]{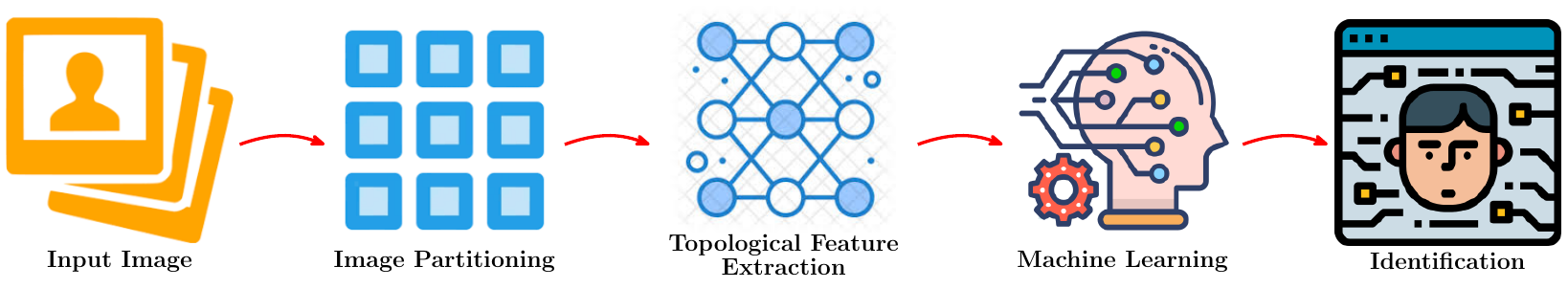}}
    \caption{Overview of the proposed topological biometric identification pipeline.}
    \label{fig:teaser}
\end{figure}

\abstract{
        \textbf{Objective} — This study introduces a novel biometric identification method based on topological invariants extracted from 2D iris images. We aim to represent the structural texture of the iris using formally defined digital homology and evaluate its classification performance.

\textbf{Methods} — Each normalized iris image $(48 \times 482 pixels)$ is partitioned into regular grids (e.g., $6 \times 54$ or $3 \times 27$). For each subregion, we compute Betti\textsubscript{0}, Betti\textsubscript{1}, and their ratio using a recently proposed algorithm for computing homology groups in 2D digital images. The resulting topological invariants form a structured feature matrix, which serves as input to traditional machine learning models including logistic regression, KNN, and SVM (with PCA and 100 randomized repetitions). For comparison, a convolutional neural network (CNN) is trained directly on the raw images.

\textbf{Results} — The logistic regression model achieved 97.78 ± 0.82\% accuracy, outperforming CNN (96.44 ± 1.32\%) and other feature-based models. The topological feature matrix exhibited high classification performance with low variance across trials.

\textbf{Conclusion} — This is the first study to use topological invariants derived from formal digital homology computations for iris-based biometric identification. The results suggest that such features provide a compact, interpretable, and highly accurate alternative to deep learning models, especially in scenarios where explainability or limited data resources are critical. Beyond iris recognition, the proposed framework holds promise for a wide range of applications, including other biometric modalities (e.g., fingerprint, face, retina), medical image analysis (e.g., tumor boundary detection, vessel topology), materials science, microscopy, remote sensing, and interpretable AI systems where topological structure carries meaningful information. These features can also be incorporated into interactive visual analytics tools to support explainable and human-in-the-loop decision-making processes. The proposed method operates efficiently on CPU-only setups, making it suitable for deployment in resource-constrained environments. Moreover, the topological features (Betti numbers and their ratios) offer interpretable and robust representations, which are particularly valuable for applications in security-critical domains.
}


    \vspace{1em}





\section{Introduction}

Biometric identification plays a critical role in modern security systems, enabling robust and individual-specific authentication. Among various biometric modalities, the iris has proven to be one of the most reliable due to its high entropy and stability over time. Traditional iris recognition approaches often rely on handcrafted features or deep learning models based on convolutional neural networks (CNNs). These models have shown excellent performance in visual classification tasks, including iris and plant cultivar recognition~\cite{zhang2021mfcis}, and histological tumor segmentation~\cite{qaiser2019fast}. However, they typically require large training datasets, long training times, and GPU-based computational resources. Furthermore, their black-box nature limits interpretability and explainability in critical applications.

Recently, topological data analysis (TDA) has emerged as a promising framework for learning from structural and geometric patterns in data. Persistent homology (PH), in particular, provides a multiscale description of shape via topological invariants such as connected components and loops~\cite{clough2020topological, songdechakraiwut2023topological}. It has been successfully applied to brain networks, tissue images, and even wafer defect classification. Nevertheless, PH-based methods require the construction of filtrations and the computation of persistence diagrams, followed by additional vectorization steps to be compatible with machine learning models. These steps can be computationally intensive and are rarely applied in biometric identification~\cite{dunaeva2016classification, assaf2025topological}.

In this study, we explore an alternative approach based on \textit{digital simplicial homology}, which allows direct computation of homology groups from 2D binary iris images. Our method extracts Betti numbers—$\beta_0$ for connected components and $\beta_1$ for loops—based on adjacency between black pixels without requiring filtrations. These features, along with their ratio $\beta_1/\beta_0$, form a compact and interpretable feature vector that reflects the intrinsic structural properties of the iris texture. They are invariant to intensity variations and small deformations, making them highly consistent and robust descriptors across different samples.

Furthermore, the flexibility of grid-based partitioning allows for the extraction of diverse topological features from multiple regions of an image. This enhances the model's adaptability and enables the generation of fine-grained feature sets without sacrificing computational efficiency. The feature space can be further enriched by combining different topological quantities depending on the problem requirements. In addition, higher-order signatures such as digital homotopy groups may be integrated in future studies to further enhance performance. To ensure mathematical correctness, we verify that the Euler characteristic calculated from Betti numbers matches the alternating sum of simplex counts.

To the best of our knowledge, this is the first study to employ formally defined digital homology groups~\cite{oztel2025} in a supervised learning framework for biometric classification. Unlike persistent homology, our approach is filtration-free, lightweight, and mathematically rigorous. It offers high classification performance, low variance, and efficient computation, making it suitable for resource-constrained environments and interpretable AI applications.

\textbf{In this work,} we present a novel topological learning pipeline for iris recognition based on digital simplicial homology. We evaluate the discriminative power of Betti-based features extracted from various grid resolutions and compare performance across logistic regression, $k$-NN, SVM, and CNN baselines. Our results demonstrate that simple yet expressive topological signatures can outperform traditional deep learning models while maintaining explainability and efficiency.

\section{Related Work}

Topological Data Analysis (TDA) has emerged as a powerful approach for extracting shape-based features from complex data, particularly in image-based classification. Persistent homology (PH), the most prominent TDA method, has been applied in various domains ranging from medical diagnostics to material science. Our study, which leverages digital homology and Betti numbers for iris recognition, builds on this foundation by proposing a filtration-free, efficient, and interpretable alternative.

Dunaeva et al.~\cite{dunaeva2016classification} applied persistent homology to the classification of endoscopy images using persistence diagrams and achieved around 90\% accuracy with a small dataset. Similarly, Adcock et al.~\cite{adcock2014classification} demonstrated the utility of topological features derived from PH in classifying hepatic lesions, especially hemangiomas, with improved robustness over geometric baselines.

Ko and Koo~\cite{ko2023novel} proposed a PH-based approach for wafer defect pattern classification that outperformed both gradient-based and CNN models. In mammographic imaging, Malek et al.~\cite{malek2024improving} enhanced the performance of traditional spatial filters by integrating persistent homology, reporting notable gains in accuracy and AUC.

In the biomedical field, Teramoto et al.~\cite{teramoto2020computer} used persistence images to identify morphological hepatocyte patterns in liver biopsies and achieved high classification accuracy. Qaiser et al.~\cite{qaiser2019fast} and Assaf et al.~\cite{assaf2025topological} utilized PH for tumor segmentation and COVID-19 detection from CT scans, respectively, with both studies reporting accuracy rates exceeding 95\%.

Beyond medical imaging, Choe and Ramanna~\cite{choe2022cubical} introduced a cubical homology-based feature extraction method, offering competitive performance on multiple image datasets. Reina-Molina et al.~\cite{reina2016effective} presented a parallelized cubical complex framework for effective homology computation, highlighting the potential of grid-based digital topologies in image processing.

From a learning perspective, Clough et al.~\cite{clough2020topological} incorporated PH into a topological loss function to guide neural networks in producing more topologically consistent segmentations. Assefa et al.~\cite{assefa2025covid} proposed a hybrid architecture combining CNNs, Transformers, and PH, achieving improved COVID-19 severity classification.

In neuroscience, Li et al.~\cite{li2020persistent} applied PH to multimodal brain networks for early-stage cognitive impairment detection, while Songdechakraiwut and Chung~\cite{songdechakraiwut2023topological} introduced a persistent loss for brain network regression and alignment.

In plant phenotyping, Zhang et al.~\cite{zhang2021mfcis} combined PH with CNNs for automatic leaf-based cultivar identification. Their hybrid model improved accuracy by leveraging topological shape features. Wang et al.~\cite{wang2025learning} proposed TSNet, a PH-based architecture with self-attention for 2D shape and graph classification, further bridging topology and deep learning.

Ehiro and Onji~\cite{ehiro2025topological} used PH for animal species classification based on leather microstructure, outperforming traditional classifiers. Fedotov et al.~\cite{fedotov2021natural} applied PH to analyze 3D structures of hydrocarbon samples, achieving robust clustering across geological conditions.

Uesugi and Ishii~\cite{uesugi2022classification} showed that PH can discriminate between amorphous and liquid states in transmission electron microscope images with over 85\% accuracy. Suresh et al.~\cite{suresh2024characterizing} used Betti numbers to quantify the evolution of feature spaces across neural network layers, correlating topological complexity with generalization ability.

Our proposed approach differs fundamentally from these works by employing digital homology rather than persistent homology. Instead of filtrations or complex vectorizations, we compute Betti numbers (connected components $\beta_0$ and loops $\beta_1$) directly from binary iris images using adjacency-based digital simplices. This reduces computational cost and preserves spatial resolution, making it particularly suitable for structured 2D biometric data such as iris images.

To the best of our knowledge, no previous study has integrated formally defined digital simplicial homology into a machine learning pipeline for biometric classification tasks. While persistent homology has dominated the literature, our approach introduces a novel and computationally efficient framework tailored for structured 2D binary images such as iris data. This makes our work the first to demonstrate the use of digital Betti numbers as topological features in a supervised classification setting.

\section{Methodology}

\begin{figure}[H]
  \centering
  \includegraphics[width=\linewidth]{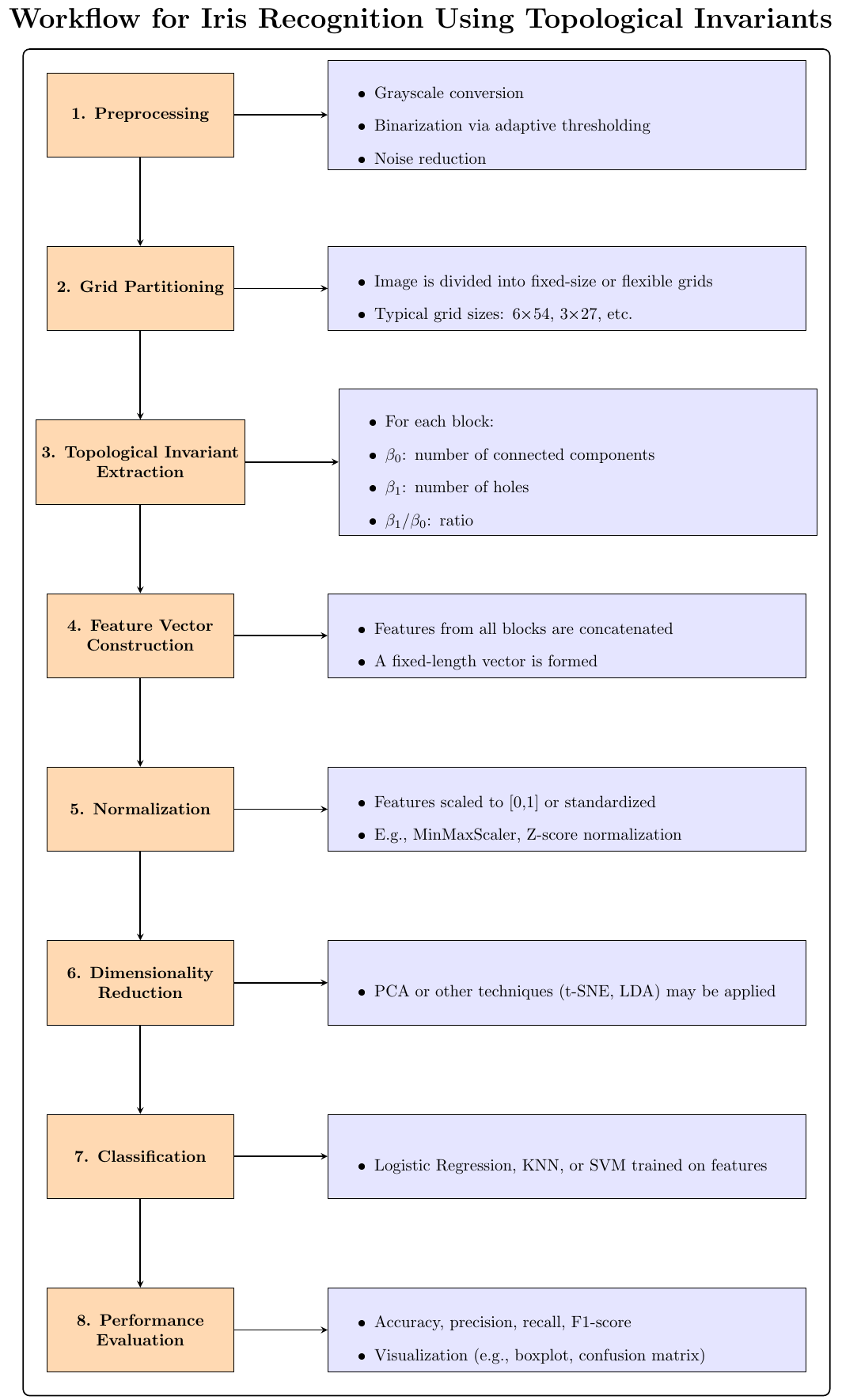}
  \caption{Workflow for iris recognition using topological invariants, including preprocessing, homology-based feature extraction, and classification.}
  \label{fig:workflow}
\end{figure}

This study proposes a digital homology-based framework for extracting topological features from 2D iris images. Our approach consists of four main stages: preprocessing and grid-based partitioning of iris images, topological feature computation using digital homology, structured feature vector construction, and supervised classification using traditional and deep learning models. The overall pipeline is summarized in Figure~\ref{fig:workflow}.

Each stage is detailed in the following subsections:

\begin{itemize}
    \item Section 3.1 describes how normalized iris images are partitioned into grid-based subregions to preserve structural details.
    \item Section 3.2 introduces the formal computation of digital homology groups, including Betti\textsubscript{0}, Betti\textsubscript{1}, and their ratio using an efficient algorithm tailored for 2D binary images.
    \item Section 3.3 explains how the computed topological features from all subregions are concatenated into a single fixed-length feature vector per image.
    \item Section 3.4 outlines the classification models and evaluation strategy, including PCA, logistic regression, SVM, KNN, and CNN.
\end{itemize}

This methodology allows the integration of mathematically interpretable topological invariants into modern pattern recognition systems, providing both accuracy and explainability in biometric identification tasks.

\subsection{Image Partitioning and Preprocessing}

The experiments in this study are conducted on the IIT Delhi Iris Image Database v1.0~\cite{iitd}, which contains 2240 near-infrared iris images from 224 subjects. Each image has a native resolution of 320×240 pixels and was captured using the JIRIS JPC1000 sensor in controlled indoor environments. From this dataset, we use five iris images per subject, totaling 1120 samples.

Rather than using raw iris images, our study operates on preprocessed, normalized iris strips that have already undergone iris segmentation, contrast enhancement, and geometric unwrapping. These rectangular iris regions are standardized at a resolution of 48×482 pixels and are widely used in prior iris recognition studies~\cite{kumar2010comparison}. Working with these ready-to-use normalized images ensures consistency and suitability for structural analysis via digital homology.

Each image is then partitioned into regular rectangular grids to enable localized topological feature extraction. The following grid configurations are employed:

\begin{itemize}
  \item \textbf{$6 \times 54$ grid:} Produces 324 subregions, enabling fine-grained structural capture.
  \item \textbf{$3 \times  27$ grid:} Produces 81 subregions, balancing resolution and computational efficiency.
  \item \textbf{$3\times 1$8 and $3 \times 9$ grids:} Used for sensitivity analysis and ablation experiments.
\end{itemize}

Since our pipeline relies on grid-based homology computation, the fixed size of 48×482 ensures that each subregion can be processed using the same algorithmic parameters. As illustrated in Figure~\ref{fig:workflow}, these partitioned subregions are the input units for digital homology-based topological feature extraction.

\subsection{Digital Homology and Topological Features}

Digital topology extends the ideas of classical topology to the discrete domain of pixel-based images. In our approach, we use formally defined homology groups based on oriented digital simplicial complexes with $8$-adjacency in $\mathbb{Z}^2$.

\paragraph*{Digital Adjacency. \cite{kong89,boxer94}} 
In $\mathbb{Z}^2$, two pixels $p = (p_1, p_2)$ and $q = (q_1, q_2)$ are said to be $c_1$-adjacent (i.e., \textbf{4-adjacent}) if they differ in exactly one coordinate and $|p_i - q_i| = 1$ for that coordinate. This corresponds to horizontal and vertical neighbors only.

They are $c_2$-adjacent (i.e., \textbf{8-adjacent}) if the number of differing coordinates is at most two and $|p_i - q_i| \leq 1$ for all $i \in \{1,2\}$. This includes diagonal neighbors as well and provides a more complete notion of connectivity for digital images.

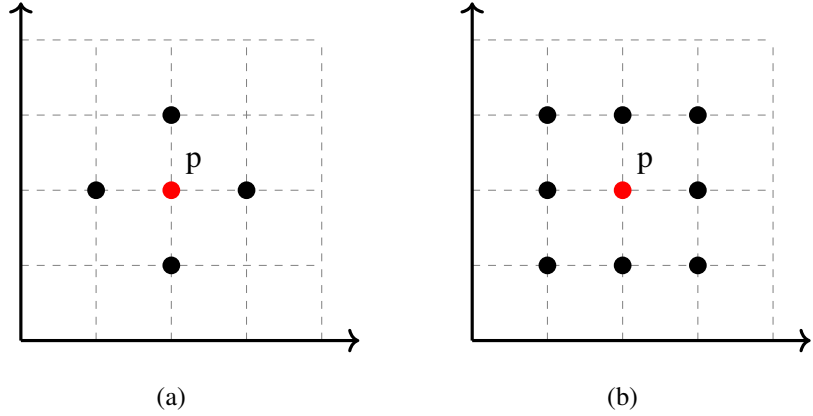
\begin{figure*}[h]
  \centering
 \begin{tikzpicture}
\draw [help lines,gray,dashed,very thin] (0,0) grid (4,4);
    \draw[->,very thick, black] (0,0) -- (4.5,0)  ;
    \draw[->,very thick, black] (0,0) -- (0,4.5)  ;
    \node at (2,3) [circle,scale=0.25mm,fill=black] {};
    \node at (2,1) [circle,scale=0.25mm,fill=black] {};
    \node at (1,2) [circle,scale=0.25mm,fill=black] {};
    \node at (3,2) [circle,scale=0.25mm,fill=black] {};

    \node at (2,2) [circle,scale=0.25mm,fill=red] {};
    \node at (2.3,2.1)[above]{\large p};
    \node at (2,-0.5)[below,black]{(a)};

    \draw [help lines,gray,dashed,very thin] (6,0) grid (10,4);
    \draw[->,very thick, black] (6,0) -- (10.5,0)  ;
    \draw[->,very thick, black] (6,0) -- (6,4.5)  ;

    \node at (8,3) [circle,scale=0.25mm,fill=black] {};
    \node at (8,1) [circle,scale=0.25mm,fill=black] {};
    \node at (9,2) [circle,scale=0.25mm,fill=black] {};
    \node at (7,2) [circle,scale=0.25mm,fill=black] {};
    \node at (9,3) [circle,scale=0.25mm,fill=black] {};
    \node at (7,1) [circle,scale=0.25mm,fill=black] {};
    \node at (9,1) [circle,scale=0.25mm,fill=black] {};
    \node at (7,3) [circle,scale=0.25mm,fill=black] {};
    \node at (8,2) [circle,scale=0.25mm,fill=red] {};
    \node at (8.3,2.1)[above]{\large p};
    \node at (8,-0.5)[below,black]{(b)};

 \end{tikzpicture}
\caption{Adjacent points of $p$ in 2D with $4$ and $8$-adjacency relations in (a) and (b), respectively.}
\label{fig:adjacency}

\end{figure*}

As shown in Figure~\ref{fig:adjacency}, 4-adjacency includes only orthogonal neighbors, while 8-adjacency captures both orthogonal and diagonal neighbors. Throughout this study, we adopt 8-adjacency for all topological computations to ensure consistency and robustness in connectedness analysis.

\paragraph*{Digital Continuity and Connectedness.}  
Let $\kappa$ be an adjacency relation defined on $\mathbb{Z}^n$. A digital image $X \subset \mathbb{Z}^n$ is said to be $\kappa$-connected if for every pair of distinct pixels $x, y \in X$, there exists a finite sequence $\{x_0, x_1, \ldots, x_r\} \subset X$ such that $x = x_0$, $y = x_r$, and each consecutive pair $(x_i, x_{i+1})$ is $\kappa$-adjacent for all $i = 0, 1, \ldots, r-1$~\cite{boxer94,ros86}. A $\kappa$-component is then defined as a maximal $\kappa$-connected subset of $X$.

Based on this notion of connectedness, continuity in the digital setting is formalized as follows. Given two digital images $X \subset \mathbb{Z}^{n_0}$ and $Y \subset \mathbb{Z}^{n_1}$ equipped with adjacency relations $\kappa_0$ and $\kappa_1$, respectively, a function $f : X \to Y$ is said to be $(\kappa_0, \kappa_1)$-continuous if the image of every $\kappa_0$-connected subset of $X$ is $\kappa_1$-connected in $Y$~\cite{boxer1999}. This forms the digital analogue of topological continuity and allows for the extension of homotopy and isomorphism concepts into digital topology.

In particular, a function $f : X \to Y$ is a $(\kappa_0, \kappa_1)$-isomorphism if it is bijective, $(\kappa_0, \kappa_1)$-continuous, and its inverse $f^{-1} : Y \to X$ is $(\kappa_1, \kappa_0)$-continuous~\cite{boxer94}. These definitions provide the foundational framework for defining homology, Euler characteristic, and other topological invariants in digital spaces.

\paragraph*{Digital Simplices.} In \cite{ars08}, the definition of a digital simplex was constructed analogously to the classical notion of a simplex in algebraic topology. However, this early formulation led to inconsistencies with topological invariants such as the Euler characteristic. A revised and consistent definition was later proposed as follows: Let $(X,\kappa)\subset \mathbb{Z}^{n}$ be a digital image and let $P = \{p_0, p_1, \ldots, p_m\} \subset X$. If for all distinct $i,j \in \{0, 1, \ldots, m\}$ the points $p_i$ and $p_j$ are $\kappa$-adjacent, then $P$ is called a digital $(\kappa,m)$-simplex, denoted by $\langle p_0, p_1, \ldots, p_m \rangle$, where $m$ is the dimension of the digital simplex \cite{boxer11}.

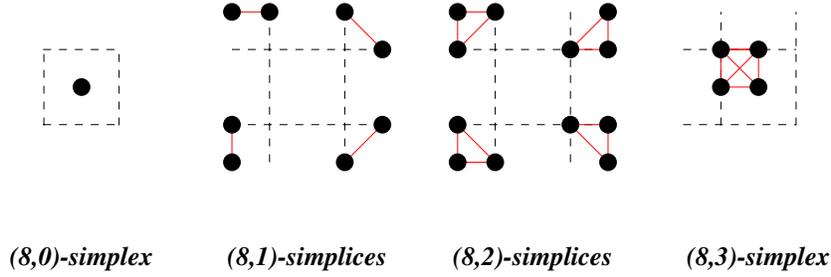
\begin{figure*}[h]
  \centering

\begin{tikzpicture}[]

\draw [very thin,dashed,gray,step=0.5] (0.99,0.99)grid(2,2);
\node at (1.5,1.5) [circle,fill=black,scale=0.25mm] {};

\draw [very thin,dashed,gray,step=0.5] (3.4999,0.5)grid(5.4999,2.5);

\draw [very thin,dashed,gray,step=0.5] (6.4999,0.5)grid(8.4999,2.5);

\draw [very thin,red](3.5,2.5)--(4,2.5);
\node at (3.5,2.5) [circle,fill=black,scale=0.25mm] {};
\node at (4,2.5) [circle,fill=black,scale=0.25mm] {};

\draw [very thin,red](3.5,0.5)--(3.5,1);
\node at (3.5,0.5) [circle,fill=black,scale=0.25mm] {};
\node at (3.5,1) [circle,fill=black,scale=0.25mm] {};

\draw [very thin,red](5,2.5)--(5.5,2);
\node at (5,2.5) [circle,fill=black,scale=0.25mm] {};
\node at (5.5,2) [circle,fill=black,scale=0.25mm] {};

\draw [very thin,red](5,0.5)--(5.5,1);
\node at (5,0.5) [circle,fill=black,scale=0.25mm] {};
\node at (5.5,1) [circle,fill=black,scale=0.25mm] {};

\draw [very thin,red](6.5,2.5)--(6.5,2)--(7,2.5)--(6.5,2.5);
\node at (6.5,2.5) [circle,fill=black,scale=0.25mm] {};
\node at (6.5,2) [circle,fill=black,scale=0.25mm] {};
\node at (7,2.5) [circle,fill=black,scale=0.25mm] {};

\draw [very thin,red](8.5,2.5)--(8.5,2)--(8,2)--(8.5,2.5);
\node at (8.5,2.5) [circle,fill=black,scale=0.25mm] {};
\node at (8.5,2) [circle,fill=black,scale=0.25mm] {};
\node at (8,2) [circle,fill=black,scale=0.25mm] {};

\draw [very thin,red](6.5,0.5)--(6.5,1)--(7,0.5)--(6.5,0.5);
\node at (6.5,0.5) [circle,fill=black,scale=0.25mm] {};
\node at (6.5,1)[circle,fill=black,scale=0.25mm] {};
\node at (7,0.5) [circle,fill=black,scale=0.25mm] {};

\draw [very thin,red](8.5,1)--(8.5,0.5)--(8,1)--(8.5,1);
\node at (8.5,1) [circle,fill=black,scale=0.25mm] {};
\node at (8.5,0.5) [circle,fill=black,scale=0.25mm] {};
\node at (8,1) [circle,fill=black,scale=0.25mm] {};

\draw [very thin,dashed,gray,step=0.5] (9.499,0.99)grid(11,2.5);
\draw [very thin,red](10,1.5)--(10,2)--(10.5,2)--(10.5,1.5)--(10,1.5);
\draw [very thin,red](10,1.5)--(10.5,2)--(10,2)--(10.5,1.5);
\node at (10,1.5) [circle,fill=black,scale=0.25mm] {};
\node at (10,2)[circle,fill=black,scale=0.25mm] {};
\node at (10.5,2) [circle,fill=black,scale=0.25mm] {};
\node at (10.5,1.5) [circle,fill=black,scale=0.25mm] {};
\node at (1.5,-0.5)[below,black]{ $\textbf{\emph{(8,0)-simplex}}$};
\node at (4.5,-0.5)[below,black]{ $\textbf{\emph{(8,1)-simplices}}$};
\node at (7.5,-0.5)[below,black]{ $\textbf{\emph{(8,2)-simplices}}$};
\node at (10.5,-0.5)[below,black]{$\textbf{\emph{(8,3)-simplex}}$};
 \end{tikzpicture}

\caption{ Digital simplices in 2Dimension with  $8-adjacency $ relations }
\label{simplicies}
\end{figure*}

Figure~\ref{simplicies} illustrates examples of $(\kappa,m)$-simplices for $m = 0,1,2,3$ in two-dimensional images with 8-adjacency. Notably, the earlier definition in \cite{ars08} failed to preserve topological properties. For instance, an $(8,3)$-simplex in 2D and a $(26,3)$-simplex in 3D are homeomorphic, yet they have different Euler characteristics (2 vs. 1), violating topological consistency. This issue has been resolved with the revised definition.

\paragraph*{Homology Groups. \cite{ars08, boxer11}} Given a chain complex
\[
0 \longrightarrow C_3^\kappa(X) \xrightarrow{\partial_3} C_2^\kappa(X) \xrightarrow{\partial_2} C_1^\kappa(X) \xrightarrow{\partial_1} C_0^\kappa(X) \longrightarrow 0,
\]
the $q$-th digital homology group is defined as
\[
H_q^\kappa(X) = \ker(\partial_q) / \operatorname{im}(\partial_{q+1}),
\]
where $\partial_q$ is the boundary operator on oriented simplices. For 2D digital images with $8$-adjacency, it has been proven that $H_q^8(X) = 0$ for all $q \geq 2$ \cite{oztel2025}. Thus, the only non-trivial topological information is encoded in $H_0^8(X)$ and $H_1^8(X)$.

\paragraph*{Betti Numbers.} The ranks of the homology groups are the digital Betti numbers:
\[
\beta_0 = \operatorname{rank}(H_0^\kappa(X)) \quad \text{and} \quad \beta_1 = \operatorname{rank}(H_1^\kappa(X)).
\]
Here, $\beta_0$ counts the number of connected components, and $\beta_1$ counts the number of holes. These values are extracted per subregion and used as features.

\paragraph*{Euler Characteristic.} The digital Euler characteristic is defined as:
\[
\chi(X, \kappa) = \sum_{q=0}^{m} (-1)^q \alpha_q(X, \kappa),
\]
where $\alpha_q$ is the number of digital $q$-simplices in $X$. In 2D, this simplifies to:
\[
\chi = \beta_0 - \beta_1.
\]
This classical invariant can serve as an additional global descriptor of image topology.

\paragraph*{Implementation.} We adopt the algorithm from \cite{oztel2025} for computing digital homology groups efficiently on binary 2D grids. It constructs digital simplicial complexes directly from pixel adjacency, applies Smith normal form to boundary matrices, and extracts Betti numbers with \(O(k^{2.1})\) time complexity, where $k$ is the number of active pixels. This approach eliminates the need for filtration or persistence steps, making it well suited for dense grid images like irises.

\begin{algorithm}[H]
\caption{Betti Numbers and Euler Characteristic via Digital Simplicial Homology}
\label{alg:betti}
\KwIn{Binary image $X \subset \mathbb{Z}^2$ with black pixels as foreground}
\KwOut{Betti numbers $\beta_0$, $\beta_1$, Euler characteristic $\chi$}

\textbf{Step 1: Simplex Construction}\;
Extract 0-simplexes: all black pixels in $X$\;
For each pair of adjacent black pixels (8-neighborhood), form a 1-simplex (edge)\;
For each triangle of 3 mutually adjacent black pixels, form a 2-simplex\;
For each tetrahedral structure of 4 mutually adjacent pixels, form a 3-simplex \;

Let $s_0$, $s_1$, $s_2$, $s_3$ be the number of $k$-simplices for $k=0,1,2,3$\;

\textbf{Step 2: Boundary Matrix Construction}\;
Build boundary matrices $B_1$, $B_2$, $B_3$ over $\mathbb{Z}$:
\Indp
$B_1$: edges $\rightarrow$ points\;
$B_2$: triangles $\rightarrow$ edges\;
$B_3$: tetrahedra $\rightarrow$ triangles\;
\Indm

\textbf{Step 3: Smith Normal Form and Betti Numbers}\;
Compute $\operatorname{rank}(B_1)$ and $\operatorname{rank}(B_2)$\;

Calculate:\;
\Indp
$\beta_0 = s_0 - \operatorname{rank}(B_1)$ \tcp*{\# of connected components}  
$\beta_1 = \operatorname{rank}(B_1) - \operatorname{rank}(B_2)$ \tcp*{\# of loops}  
\Indm

\textbf{Step 4: Euler Characteristic Calculation}\;

\vspace{0.5ex}

$\chi = \beta_0 - \beta_1 = s_0 - s_1 + s_2 - s_3$

\textbf{Return:} $\beta_0$, $\beta_1$, $\chi$
\end{algorithm}

\vspace{1ex}
To improve reproducibility and clarify the digital homology computation process, Algorithm~\ref{alg:betti} outlines the key steps used to extract Betti numbers from binary images using adjacency-based digital simplicial complexes. This procedure is based on the method proposed by Öztel et al.~\cite{oztel2025}. To validate the correctness of homology computations, the Euler characteristic is calculated both from Betti numbers and from the alternating sum of simplex counts. In 2D digital images, $\beta_2$ and $\beta_3$ are expected to be zero, and consistency between the two results provides a strong correctness check.

To promote reproducibility and facilitate further research, the Python implementation of the algorithm for computing digital homology groups (DHGComp) is publicly available\footnote{\url{https://github.com/ahmetoztel/DHGComp/tree/main}}. This repository contains the full source code used in this study for calculating Betti numbers and Euler characteristics from digital images, as well as examples and documentation. Although the core implementation was previously shared in a related publication, we include it here for completeness and accessibility.

\subsection{Feature Vector Construction}

In this study, each iris image is represented by a set of topological features extracted from multiple local regions. Given a normalized iris image of size $48 \times 482$, we divide the image into various spatial grids, such as $3 \times 27$, $3 \times 18$, or $6 \times 54$, resulting in multiple non-overlapping rectangular subregions. For each subregion, three topological signatures are computed: Betti$_0$ (the number of connected components), Betti$_1$ (the number of one-dimensional holes), and their ratio $\beta_1 / \beta_0$. These quantities are calculated based on the theory of digital homology groups defined over $(\mathbb{Z}^2, 8)$ with 8-adjacency, as detailed in the previous section.

Let $G$ denote the number of subregions obtained by a grid partitioning. Then, each iris image is transformed into a topological feature vector of length $3G$. For example, with a $3 \times 27$ grid, the number of subregions is $81$, resulting in a 243-dimensional feature vector. This representation captures both local and global topological complexity of the iris texture while remaining invariant under minor deformations.

The motivation for using such topological features is twofold. First, Betti numbers provide a robust summary of structural patterns in binary images that is resilient to noise and deformation. Second, computing these features from localized regions enhances discriminability by capturing spatial variations across the iris. As a result, the final feature vector serves as a discriminative and compact representation suitable for downstream classification tasks.

\subsection{Machine Learning Framework}

To evaluate the discriminative capacity of the extracted topological features, we designed a supervised multi-class classification framework using Logistic Regression. The task was to identify individuals based on their iris image features, where each class corresponds to a unique subject.

We used a dataset consisting of normalized iris images from 224 individuals, each represented by five images, resulting in 1120 samples. The input features include 102 topological descriptors (Betti numbers and their ratios) computed over 34 spatial partitions per image. Labels were derived from the filenames by parsing person identifiers.

All features were scaled using min-max normalization. Dimensionality reduction was performed with Principal Component Analysis (PCA), preserving 99\% of the total variance. Importantly, both normalization and PCA were fit only on the training data and applied to the test data to avoid information leakage.

We used a Logistic Regression classifier with $C=10$ and $L_2$ regularization, trained via the LBFGS solver. The performance was evaluated using 10 independent stratified train-test splits (80\%/20\%) with varying random seeds. Accuracy scores were computed for each run and summarized using descriptive statistics.

All experiments were implemented in Python using the \texttt{scikit-learn}, \texttt{numpy}, \texttt{pandas}, and \texttt{matplotlib} libraries. The entire pipeline ensures repeatability and statistical robustness. The Python implementation of our logistic regression-based classification is publicly available on GitHub.\footnote{\url{https://github.com/ahmetoztel/Iris_Top_Log_Reg}}

\vspace{1em}
\noindent
In the following section, we report the experimental results, including average accuracy, standard deviation, and comparative visualizations.

\section{Results}

This section presents a comprehensive evaluation of the proposed topological feature-based biometric identification framework. We report classification accuracies across multiple machine learning models, grid partitioning strategies, and compare the performance with a CNN-based approach. All experiments were conducted on a standard desktop computer with an Intel Core i7 CPU, 16GB RAM, and no GPU acceleration, using Python 3.10 and scikit-learn. All codes and results are openly available to support reproducibility.

\subsection{Logistic Regression Performance}

\begin{figure}[H]
\centering
\includegraphics[width=0.8\linewidth]{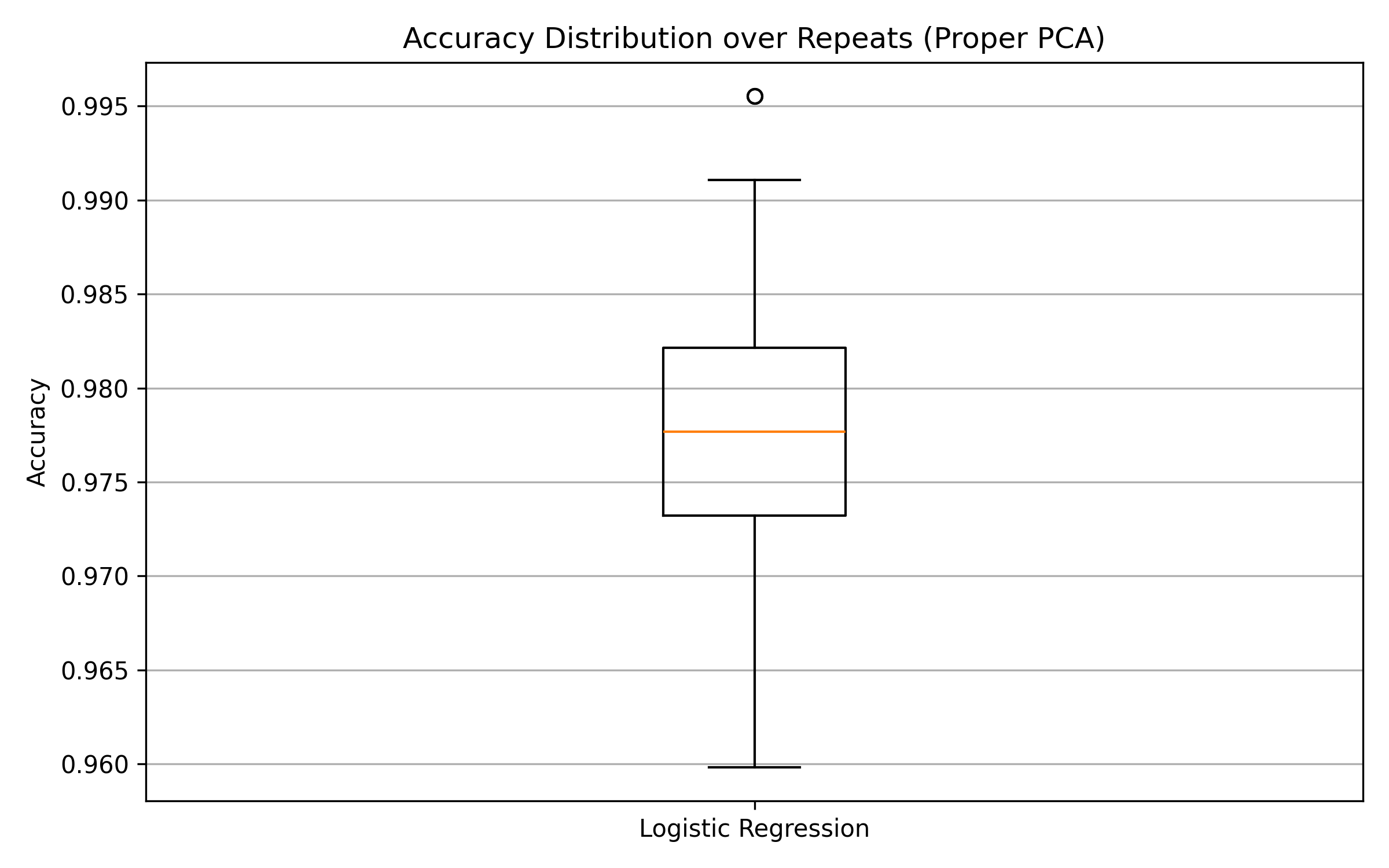}
\caption{Accuracy distribution of Logistic Regression over 100 runs using $6\times54$ topological features.}
\label{fig:logreg_boxplot_corrected}
\end{figure}

The Logistic Regression (LR) model was tested over 100 independent runs using a fixed $6\times54$ grid-based topological descriptor. As shown in Figure~\ref{fig:logreg_boxplot_corrected}, the model achieved a mean accuracy of $97.78\%$ with a low standard deviation of $0.82\%$, indicating both high performance and robustness.

In addition to numerical evaluation, we also visualized the spatial distribution of topological features across the grid, as shown in Figure~\ref{fig:topo_maps}.

\subsection{Topological Feature Visualization}

\begin{figure}[H]
  \centering
  \includegraphics[width=\linewidth]{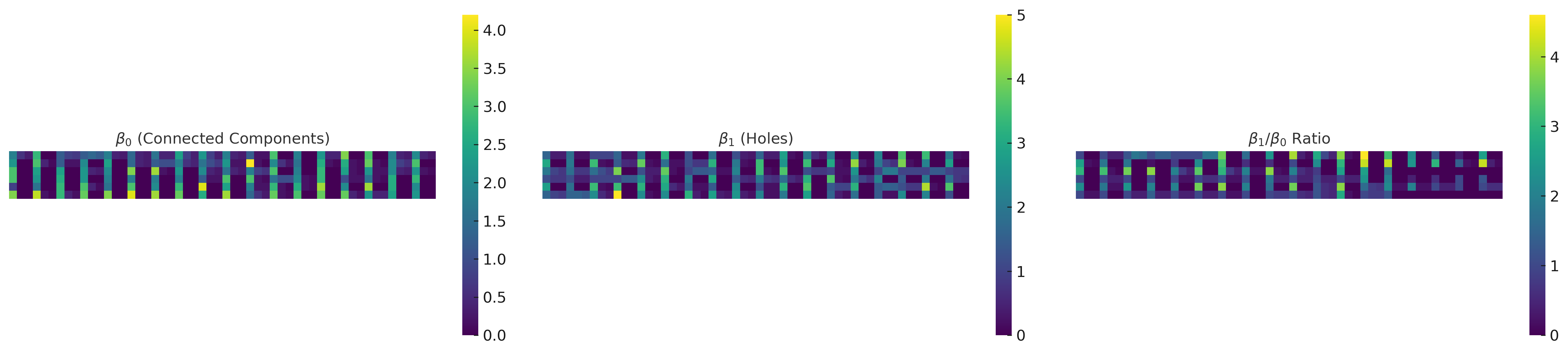}
  \caption{Topological feature distributions over the $6 \times 54$ grid for a sample subject: $\beta_0$ (connected components), $\beta_1$ (holes), and the ratio $\beta_1/\beta_0$. These spatial maps highlight the heterogeneous complexity across iris regions.}
  \label{fig:topo_maps}
\end{figure}

To better understand the spatial distribution of the extracted topological features, we computed the mean $\beta_0$, $\beta_1$, and $\beta_1/\beta_0$ values across all five images of a randomly selected subject. These values were mapped back to the original $6 \times 54$ grid structure and visualized as heatmaps (Figure~\ref{fig:topo_maps}).

The resulting distributions reveal distinct patterns across iris regions. In particular, the $\beta_1/\beta_0$ ratio highlights regions with disproportionately high loop complexity, indicating areas of increased texture or structural variability. Such visualizations can provide additional insight into how topology-based features encode biometric identity.

\begin{figure}[H]
  \centering
  \includegraphics[width=\linewidth]{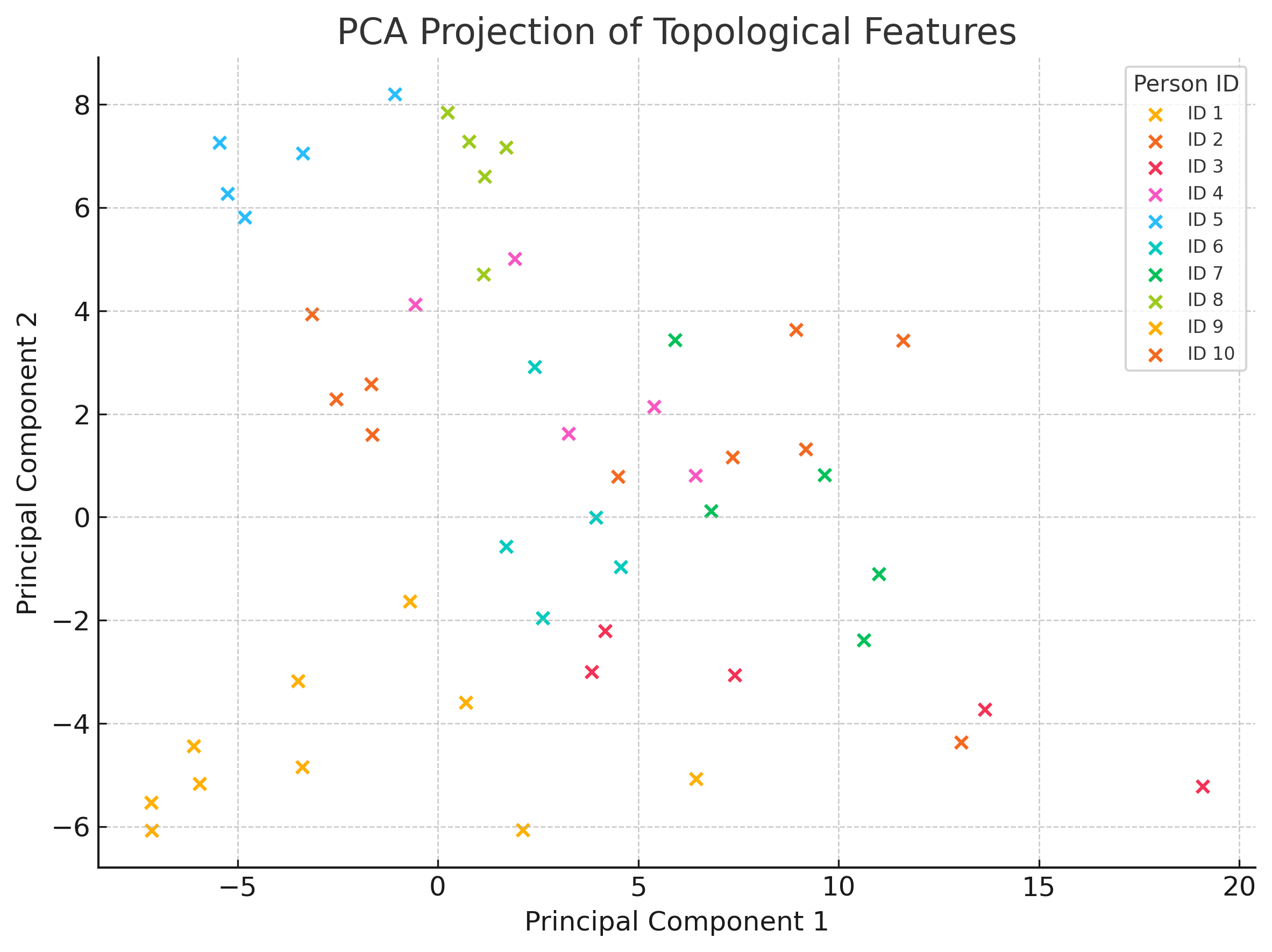}
  \caption{2D PCA projection of topological features for the first 10 individuals. The separation indicates strong discriminative capacity of the Betti-based descriptors.}
  \label{fig:pca_projection}
\end{figure}

To further explore the discriminative power of the topological feature space, we performed Principal Component Analysis (PCA) on the extracted features. Figure~\ref{fig:pca_projection} visualizes the PCA projection of the topological features for a representative subset of 10 individuals. This limited selection allows for clear visual separation and class-level interpretation.

The resulting clusters reveal that the proposed features capture meaningful structural distinctions across different individuals, as points corresponding to the same subject tend to group together in the reduced space. While visualization for all 224 subjects is possible, it would hinder interpretability due to overlapping data points and color ambiguity. Therefore, we opted to illustrate the clustering behavior on a manageable subset for clarity.

These results reinforce the notion that Betti-based topological signatures form a compact and separable embedding space suitable for biometric classification.

\subsection{Comparison with Other Classifiers}

\begin{figure}[H]
\centering
\includegraphics[width=0.8\linewidth]{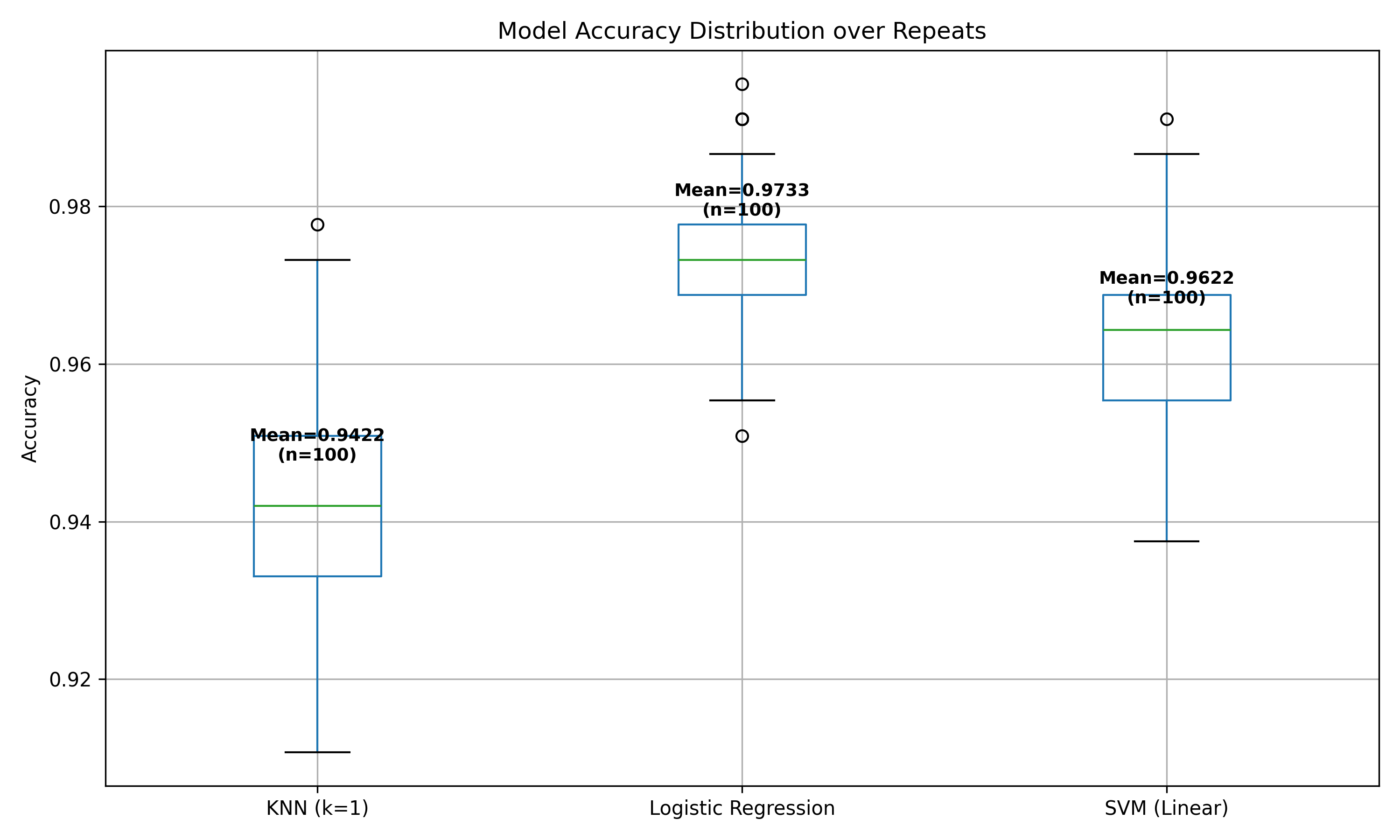}
\caption{Boxplot comparison of classification accuracies for three models.}
\label{fig:model_boxplot}
\end{figure}

To benchmark the LR performance, we also evaluated K-Nearest Neighbors (KNN, $k=1$) and Support Vector Machine (SVM, linear kernel) models using the same feature set. As summarized in Table~\ref{tab:model_comparison} and visualized in Figure~\ref{fig:model_boxplot}, LR outperformed both KNN and SVM in terms of average accuracy and variability.

\begin{table}[H]
\centering
\begin{tabular}{lccc}
\toprule
\textbf{Model} & \textbf{Mean} & \textbf{Std} & \textbf{Mean $\pm$ Std} \\
\midrule
KNN ($k=1$) & 0.9422 & 0.0138 & $0.9422 \pm 0.0138$ \\
SVM (Linear) & 0.9622 & 0.0108 & $0.9622 \pm 0.0108$ \\
Logistic Regression & 0.9733 & 0.0090 & $0.9733 \pm 0.0090$ \\
\bottomrule
\end{tabular}
\caption{Comparison of different classifiers over 100 runs using $6\times54$ grid features.}
\label{tab:model_comparison}
\end{table}

\subsection{CNN Baseline Comparison}

To assess the efficacy of topological features against visual-based representations, we compared the LR model with a Convolutional Neural Network (CNN) trained on the same iris dataset. The CNN model consists of two convolutional layers with ReLU activations and max-pooling, followed by two fully connected layers with ReLU and softmax outputs. The input images were resized to $64 \times 64$ resolution. The model was trained for 20 epochs using the Adam optimizer with a learning rate of 0.001 and cross-entropy loss. In each of the 100 repeated experiments, the dataset was split into 80\% training and 20\% test sets with stratification by person ID.

  As shown in Table~\ref{tab:cnn_comparison}, the CNN achieved a mean accuracy of $96.44\%$, which is lower than LR's $97.78\%$, indicating that topological signatures can offer superior performance in certain biometric tasks.

\begin{table}[H]
\centering
\begin{tabular}{lc}
\toprule
\textbf{Model} & \textbf{Mean $\pm$ Std} \\
\midrule
CNN & $0.9644 \pm 0.0132$ \\
\textbf{Logistic Regression} & $\mathbf{0.9778 \pm 0.0082}$ \\
\bottomrule
\end{tabular}
\caption{CNN vs Logistic Regression performance over 100 runs.}
\label{tab:cnn_comparison}
\end{table}

\begin{figure}[H]
\centering
\includegraphics[width=0.8\linewidth]{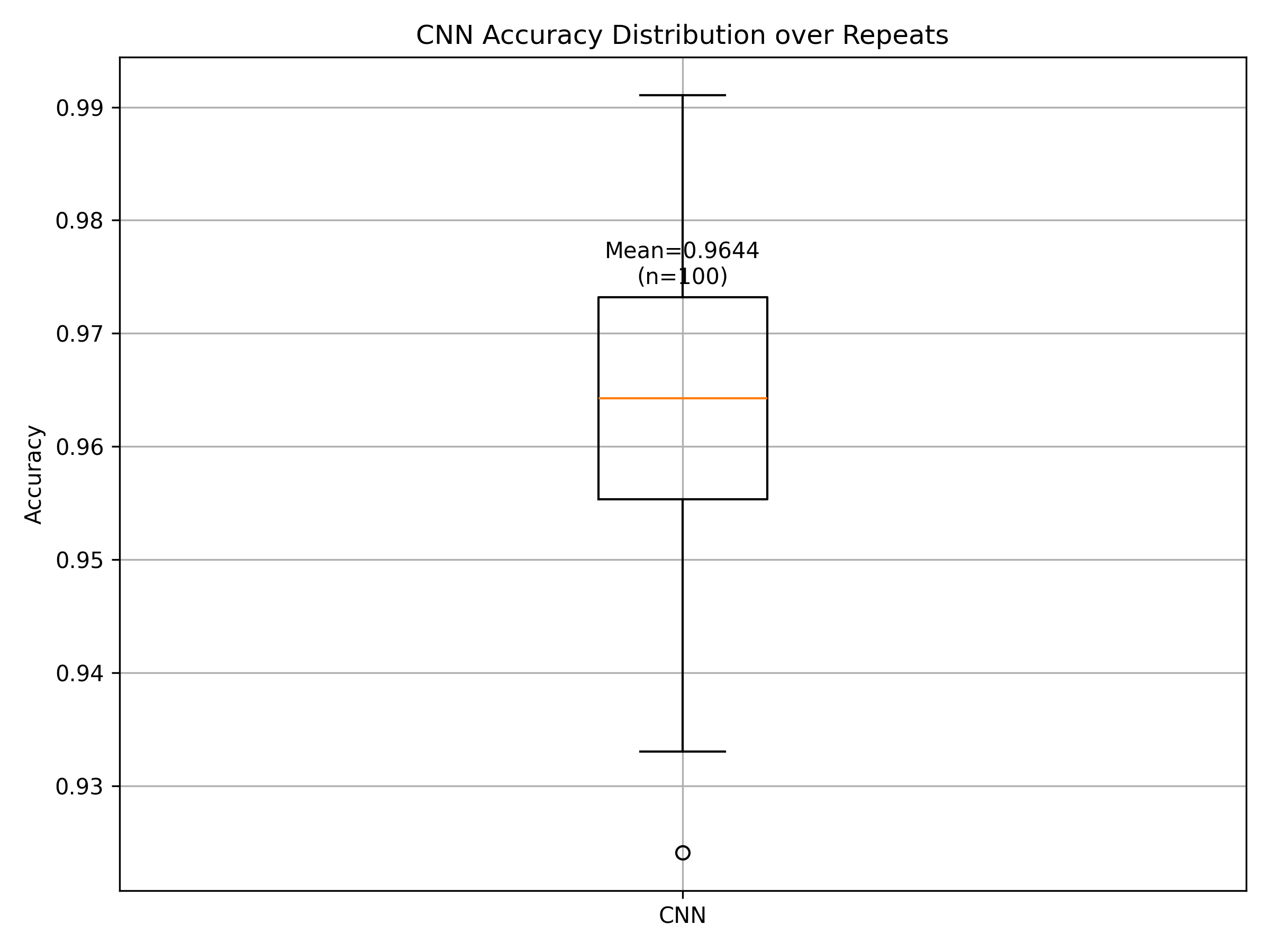}
\caption{Accuracy distribution of CNN over 100 runs.}
\label{fig:cnn_boxplot}
\end{figure}

\subsection{Effect of Grid Partitioning}

\begin{figure}[H]
\centering
\includegraphics[width=0.8\linewidth]{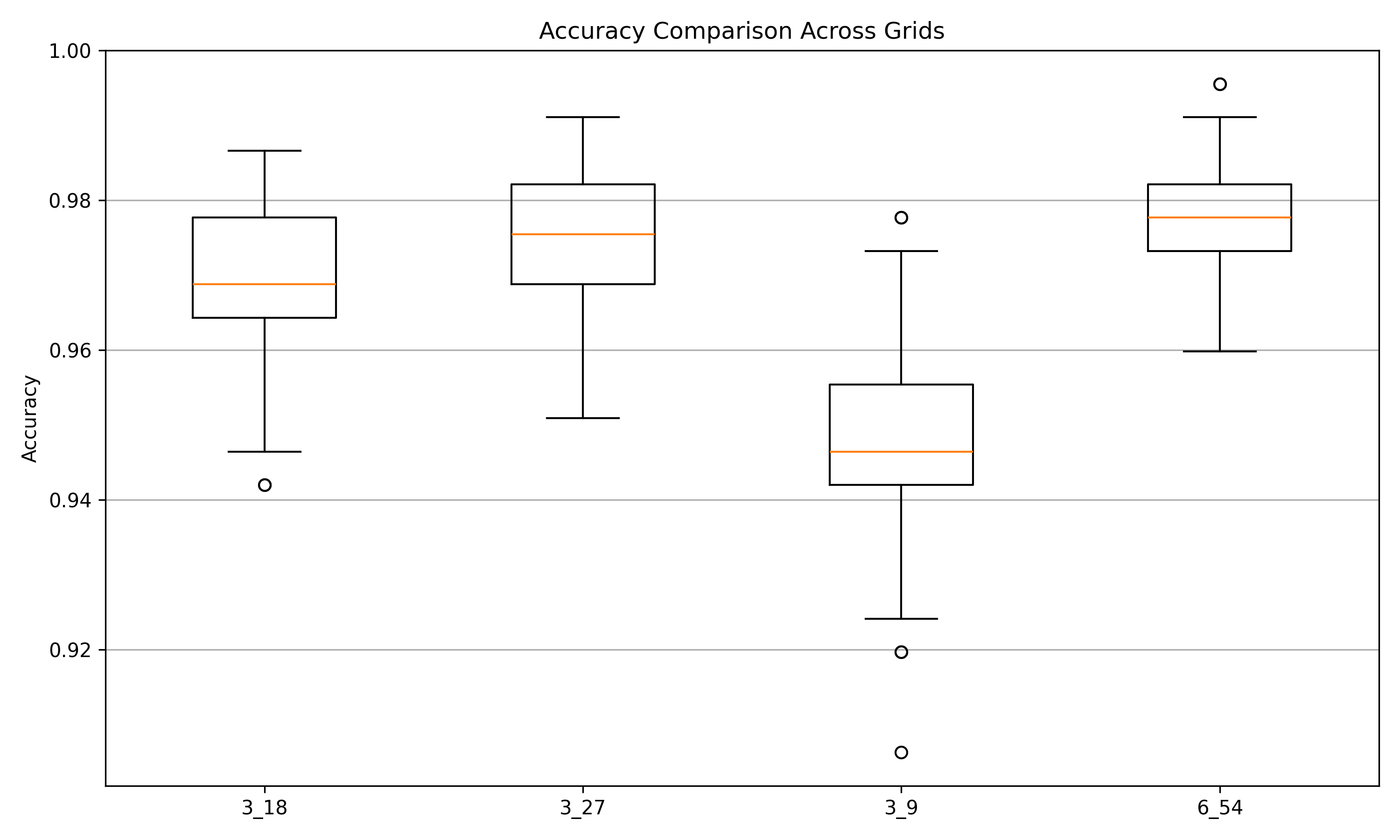}
\caption{Effect of grid resolution on classification accuracy using Logistic Regression.}
\label{fig:grid_boxplot}
\end{figure}

We further investigated how different grid partition strategies affect classification accuracy. As shown in Table~\ref{tab:grid_effect} and Figure~\ref{fig:grid_boxplot}, finer partitions (e.g., $6\times54$) lead to better performance, while coarser grids reduce discriminative power.

\begin{table}[H]
\centering
\begin{tabular}{lccc}
\toprule
\textbf{Grid} & \textbf{Mean} & \textbf{Std} & \textbf{Mean $\pm$ Std} \\
\midrule
$6\times54$ & 0.9778 & 0.0082 & $0.9778 \pm 0.0082$ \\
$3\times27$ & 0.9741 & 0.0097 & $0.9741 \pm 0.0097$ \\
$3\times18$ & 0.9688 & 0.0104 & $0.9688 \pm 0.0104$ \\
$3\times9$ & 0.9463 & 0.0126 & $0.9463 \pm 0.0126$ \\
\bottomrule
\end{tabular}
\caption{Accuracy of Logistic Regression with varying grid sizes.}
\label{tab:grid_effect}
\end{table}

\section{Discussion}

The experimental results demonstrate that topological features, particularly Betti numbers and their ratios derived from grid-based partitioning of normalized iris images, can be effectively leveraged for biometric identification. The best performance was achieved using a $6 \times 54$ grid with logistic regression, yielding a mean accuracy of $97.78\% \pm 0.82\%$ over 100 repeated trials. This outperformed both traditional machine learning methods such as $k$-NN and SVM, and a baseline CNN model trained directly on image pixels.

Compared to deep learning-based approaches, the topological pipeline offers two main advantages. First, it achieves higher accuracy with lower variance, indicating improved stability. Second, it requires significantly fewer computational resources. While CNN models typically demand GPU acceleration and longer training times, our topology-based pipeline is lightweight and interpretable, operating efficiently on standard CPU hardware.

The results also indicate that finer partitioning (i.e., more granular grid sizes) generally improves the model performance. However, this comes at the cost of increased feature dimensionality. Fortunately, the application of PCA preserved $99\%$ of the variance and mitigated overfitting, enabling robust classification.

Our findings are consistent with earlier research on the use of topological features in image-based classification tasks. Prior studies have demonstrated the utility of persistent homology (PH) in domains such as medical image segmentation~\cite{qaiser2019fast, teramoto2020computer}, COVID-19 detection~\cite{assaf2025topological, assefa2025covid}, wafer defect recognition~\cite{ko2023novel}, and plant identification~\cite{zhang2021mfcis}. These studies highlight PH’s robustness to noise and deformations, and its ability to capture multi-scale geometric structures.

However, PH-based approaches typically rely on filtrations and require persistence diagrams to be vectorized before they can be used in machine learning pipelines, which increases computational complexity and reduces interpretability. By contrast, our method computes Betti numbers directly using formally defined digital simplicial homology without the need for filtrations. This not only reduces computational burden but also preserves the spatial granularity of 2D grid images such as iris textures.

To the best of our knowledge, this is the first study to integrate digital homology---as rigorously defined in digital topology~\cite{oztel2025}---into a supervised learning framework for biometric identification. While cubical homology has previously been explored for general image classification~\cite{choe2022cubical, reina2016effective}, its application to biometrics remains unexplored. Our approach demonstrates that digital homology-based features can offer competitive or superior performance compared to deep learning methods, especially in structured visual domains where spatial consistency is crucial.

Moreover, our findings are in line with recent studies showing the relevance of topological invariants such as Betti numbers for interpreting neural networks and embedding spaces~\cite{suresh2024characterizing}. These results further validate that topological summaries are not only theoretically sound but also practically meaningful in complex pattern recognition tasks.

Recent studies continue to highlight the versatility and robustness of topological data analysis (TDA) in diverse applications such as biomedicine, security, and real-time data processing. For instance, Liu et al.~\cite{liu2024topological} proposed a gait-based biometric recognition method using wearable IMU sensors, where topological descriptors extracted from phase-space trajectories achieved over $96\%$ accuracy, even under varying sensor placements and terrain conditions. In the biomedical domain, Rostami et al.~\cite{rostami2025topological} employed TDA on breast cancer gene expression data to uncover a previously unidentified HER-2 positive luminal B subtype, demonstrating the explanatory power of topological structures in high-dimensional biological datasets. Complementarily, Luo et al.~\cite{luo2024efficient} developed a topology-driven clustering framework, Tri-Squeezing SOM for Clustering (TSSC), that effectively handles imbalance-drifting in streaming biomedical data through self-organizing maps and nested topological representations.

These studies collectively underscore the promise of topological features in constructing efficient, explainable, and adaptable learning systems. In contrast to persistent homology or self-organizing models, our framework directly computes digital Betti numbers on 2D image grids without requiring filtrations or post-hoc vectorizations. By leveraging the formalism of digital homology~\cite{oztel2025}, our work introduces a lightweight, interpretable, and effective biometric recognition pipeline rooted in discrete algebraic topology.

The modularity of our pipeline enables potential integration into a visual analytics interface where users could interactively adjust grid resolution, visualize region-wise topological features, and monitor model response in real-time.

While the proposed method achieves high identification accuracy with low variance, it should be noted that the CNN architecture used in the comparative analysis is intentionally kept simple. This decision was made to highlight the performance of the proposed topological features in a clear and interpretable manner without the influence of deeper and more complex network-specific optimizations. As this study serves as a foundational and exploratory work, the goal was to ensure transparency and methodological clarity rather than to optimize CNN performance. Nevertheless, future studies may consider evaluating the integration of topological features with more advanced architectures such as ResNet, MobileNet, or EfficientNet to further assess the scalability and complementary strengths of this approach.

An important factor contributing to the high classification performance is the nature of the selected topological features themselves. The number of connected components ($\beta_0$), holes ($\beta_1$), and their ratio $\beta_1 / \beta_0$ are not merely numerical summaries; they encode intrinsic structural properties of the iris texture. These descriptors are invariant under pixel-level intensity changes and are largely unaffected by small deformations, making them robust across different samples. Moreover, because they reflect the connectivity and geometric complexity of image regions, they serve as consistent and valid representations with high discriminatory capacity. This structural invariance is especially advantageous in biometric tasks, where preserving identity-specific patterns while ignoring irrelevant noise is critical.

Nonetheless, there are several limitations. First, the current study uses only the IITD iris dataset with pre-normalized images; thus, the generalizability of our approach to raw or noisy images remains to be tested. Second, while the current pipeline focuses on binary topological features, future work may explore persistent homology and other advanced invariants to capture multi-scale structures more effectively.

As a preliminary and exploratory study, our primary aim is to introduce and validate the feasibility of using digital homology features for biometric recognition. By leveraging interpretable topological features in a lightweight learning pipeline, this study contributes to the integration of topological data analysis into practical, explainable, and visually traceable machine learning applications, aligning with the goals of topological visualization frameworks.

In conclusion, this study supports the integration of algebraic topology and machine learning as a promising direction for interpretable and efficient visual recognition systems. Future research may include combining topological features with deep embeddings, extending to other biometric modalities, and evaluating robustness under image distortions and adversarial conditions.

\section{Conclusion}

This study proposes a novel and computationally efficient approach to biometric identification by leveraging digital simplicial homology. Unlike persistent homology-based methods, the proposed technique directly extracts topological features—specifically Betti numbers—from 2D binary iris images without requiring filtrations or persistence diagram vectorizations. This results in a fast, interpretable, and reproducible pipeline that is particularly suitable for structured biometric data.

Experimental results show that the topological features derived from fine-grained grid partitions, when combined with logistic regression and PCA, achieve a mean accuracy of $97.78\% \pm 0.82\%$, outperforming both classical machine learning models and baseline CNNs. In addition to accuracy, the method offers several important advantages:

\begin{itemize}
  \item It directly produces fixed-length feature vectors, which can be used seamlessly in any machine learning pipeline.
  \item Grid partitioning flexibility enables the generation of multiple feature sets from the same image, enhancing the adaptability of the model to the task.
  \item The feature vector can be easily extended by including different combinations of topological invariants, depending on the problem context.
  \item The topological features used in this study—namely the number of connected components ($\beta_0$), loops ($\beta_1$), and their ratios—are inherently structural properties of the images. As such, they provide consistent, valid, and highly representative signatures across samples.
  \item It opens the door to future integration of higher-order topological signatures such as digital homotopy groups, which may further improve classification performance.
  \item The Euler characteristic is computed through both Betti numbers and simplex counts, ensuring mathematical consistency and correctness of the topological extraction.
\end{itemize}

The topological descriptors used—$\beta_0$, $\beta_1$, and their ratios—encode structural characteristics that are intrinsic to the shape and connectivity of iris textures. Their consistency across images, independence from pixel intensity variations, and abstract nature grant them strong representational power and robustness, which directly contribute to the high classification accuracy observed.

To the best of our knowledge, this is the first study to integrate formally defined digital homology groups into a supervised learning framework for biometric identification. This positions digital topological analysis as a promising alternative or complement to deep learning, particularly in scenarios where computational resources are limited or model interpretability is desired.

Future work will investigate robustness to noise, applications to other biometric modalities, and the combination of topological features with learned visual embeddings to build hybrid models.

\bibliographystyle{abbrv-doi}
\bibliography{refs}

\end{document}